\documentclass[10pt,letterpaper]{article}
\usepackage[top=0.85in,footskip=0.75in]{geometry}
\pdfoutput=1
\usepackage[utf8]{inputenc} 
\usepackage[T1]{fontenc}    
\usepackage{hyperref}       
\usepackage{url}            
\usepackage{booktabs}       
\usepackage{amsfonts}       
\usepackage{nicefrac}       
\usepackage{microtype}      
\usepackage{todonotes}
\usepackage{amsmath,amssymb}
\usepackage{dsfont}
\usepackage{textgreek}

\usepackage[aboveskip=1pt,labelfont=bf,labelsep=period,justification=raggedright,singlelinecheck=off]{caption}
\makeatletter
\renewcommand{\@biblabel}[1]{\quad#1.}
\makeatother
\reversemarginpar

\usepackage{lastpage,fancyhdr,graphicx}
\usepackage{epstopdf}
\DeclareFontEncoding{LGR}{}{}
\DeclareTextSymbol{\~}{LGR}{126}
\usepackage{subfig}

\begin{document}
\vspace*{0.35in}

\begin{flushleft}
{\Large
\textbf\newline{The Kernel Mixture Network: A Nonparametric Method for Conditional Density Estimation of Continuous Random Variables}
}
\newline
\\
Luca Ambrogioni\textsuperscript{1}, Umut Güçlü\textsuperscript{1}, Marcel van Gerven\textsuperscript{1} and
Eric Maris\textsuperscript{1}
\\
\bigskip
\bf{1} Radboud University, Donders Institute for Brain, Cognition and Behaviour, Nijmegen, The Netherlands 
\\
\bigskip
* l.ambrogioni@donders.ru.nl

\end{flushleft}

\section*{Abstract}
This paper introduces the kernel mixture network, a new method for nonparametric estimation of conditional probability densities using neural networks. We model arbitrarily complex conditional densities as linear combinations of a family of kernel functions centered at a subset of training points. The weights are determined by the outer layer of a deep neural network, trained by minimizing the negative log likelihood. This generalizes the popular quantized softmax approach, which can be seen as a kernel mixture network with square and non-overlapping kernels. We test the performance of our method on two important applications, namely Bayesian filtering and generative modeling. In the Bayesian filtering example, we show that the method can be used to filter complex nonlinear and non-Gaussian signals defined on manifolds. The resulting kernel mixture network filter outperforms both the quantized softmax filter and the extended Kalman filter in terms of model likelihood. Finally, our experiments on generative models show that, given the same architecture, the kernel mixture network leads to higher test set likelihood, less overfitting and more diversified and realistic generated samples than the quantized softmax approach.

\section{Introduction}
Almost all probabilistic machine learning problems can be interpreted in terms of conditional density estimation (CDE). For example, in Bayesian filtering the aim is to estimate the probability density of the current state of a dynamical system given a series of indirect and noise-corrupted past observations \cite{sarkka2013bayesian}. Another popular example is generative modeling, where realistic synthetic data such as images and sounds are generated by expressing their joint distribution as a product of univariate conditional distributions \cite{oord2016wavenet, van2016conditional}. Deep neural networks (DNN) are powerful tools for approximating extremely complex functional dependencies. It is therefore natural to use DNN for solving CDE problems. When the conditional distribution is defined over a finite set, this can be done using a softmax distribution in the outer layer. Two main strategies have been used in the continuous case: mixture density networks and quantized softmax networks. In the mixture approach the conditional density is approximated as a convex mixture of simple density functions whose parameters are determined by the output of a DNN \cite{bishop1994mixture, theis2015generative, salimans2017pixelcnn}. Conversely, in a quantized softmax network, the continuous variable is discretized into a finite number of bins that are then treated as discrete classes \cite{oord2016pixel}. This strategy requires a careful design of the quantization scheme and results in very sparse gradients since every training example provides strong error signals for a single bin. In other words, the softmax network does not exploit the topology of the real numbers in the backpropagation of the error signals. Despite these shortcomings, the quantized approach is often preferred over mixture density networks \cite{oord2016pixel, oord2016wavenet}. The main reason for this choice is that the softmax distribution can approximate arbitrarily complex conditional densities, as it does not make any parametric assumptions. Fortunately, in the statistics literature there are several examples of truly nonparametric density estimation methods that do not rely on a quantization scheme and can therefore exploit the continuous nature of random variables \cite{rosenblatt1956remarks,parzen1962estimation}. Drawing inspiration from these techniques, we introduce a new CDE method, the \emph{kernel mixture network} (KMN), which combines the flexibility of the quantized approach with the benefits of a continuous modeling of conditional densities. The main idea is to model the conditional density as a linear combination of kernel functions centered on a subset of training points, where the weights are determined by the output of a DNN. While we will focus on conditional densities defined over either the real numbers or the unitary circle, the kernel functions can potentially be defined over arbitrary manifolds and even discrete topological objects such as graphs. 

We validate the KMN on two important applications: Bayesian filtering and generative modeling. In the Bayesian filtering example, we show that the KMN can be used together with a CNN architecture to construct new powerful approximate Bayesian filters that can be applied to arbitrarily nonlinear dynamical systems and recover complex non-Gaussian posterior densities. To the best of our knowledge we are the first to introduce this family of CNN-based Bayesian filters, closely related to the recently introduced ConvNet smoother \cite{ambrogioni2017estimating}. As a second application, we use the KMN for constructing a probabilistic generative model, based on LSTM units, that learns the conditional density of each principal component of the training set given all the higher variance components. We used this generative network in order to generate realistic images of human faces. The generative approach, which we named LSTM-PCA, is original by itself and is an alternative to more computationally expensive methods such as pixelRNN \cite{van2016conditional}. In our analysis we show that, given the same LSTM-PCA architecture, the KMN approach significantly outperforms the quantized softmax approach in terms of both model likelihood and realism and variability of the generated images.

\subsection{Related work}
The KMN is related to several other deep learning methods. The output of a KMN is expressed as a linear combination of kernel functions centered at the training points. This feature is shared with kernel methods such as Gaussian process regression and support vector machines. Kernel methods have recently been combined with DNNs. These hybrid approaches exploit the representational power of DNNs in order to construct complex kernel functions \cite{wilson2016deep, wilson2016stochastic}. Our method differs from these approaches in that we do not learn the kernel functions. Instead we use a DNN in order to determine the mixing weights of a fixed set of kernels. Furthermore, our kernels do not need to be positive semi-definite and the KMN can be trained using standard gradient descent, without resorting to stochastic variational learning. 

The functional form of the KMN output is similar to radial basis function (RBF) networks \cite{buhmann2003radial}. However, in contrast to most RBF methods, the KMN method 1) does not use radial activation functions in the hidden layers, 2) does not require training of the center points, 3) is not limited to radially symmetric kernel functions, 4) uses the negative log likelihood as loss function and 5) uses a whole family of kernels for each center point instead of a single kernel. 

The KMN is complementary to the recently introduced geometric deep learning methods, which generalize convolutional architectures to arbitrary input manifolds \cite{monti2016geometric, bronstein2016geometric}. In fact, the output of a KMN can be defined on any arbitrary output manifold by appropriately choosing the kernel functions. Therefore, geometric deep learning and KMN can be combined for constructing deep convolutional mappings between manifolds. 

Our Bayesian filtering approach can also be interpreted as a new application of \textepsilon-free approximate Bayesian inference \cite{papamakarios2016fast}. It differs from other deep filtering methods because it approximates the posterior distribution from a series of synthetic samples instead of relying on a variational Bayes scheme \cite{krishnan2015deep, archer2015black}. 

We formulate our generative model by modifying existing approaches that rely on autoregressive models such as CharRNN \cite{charrnn}, PixelCNN \cite{van2016conditional}, PixelRNN \cite{pixelrnn}, WaveNet \cite{oord2016wavenet} and ByteNet \cite{kalchbrenner2016neural}. These generative models use either recurrent or convolutional neural networks to model the conditional densities of individual variables given all the previous variables by using the quantized softmax loss function. The procedure can be implemented in a relatively simple manner for one-dimensional sequential data with an inherent ordering. On the other hand, applying these methods on data that lack inherent ordering is rather complex. For example, PixelRNN and PixelCNN resort to the use of multiple streams that process images horizontally and vertically as well as complicated masking to generate an RGB image pixel by pixel while ensuring valid conditional densities. We propose a simple solution to the problem of generating RGB images with autoregressive models that relies on principle component analysis. Our LSTM-PCA approach automatically exploits the heterogeneous variance of the components and also enables faster training and sampling by only considering the high variance components. 

\section{Background}

A CDE problem consists of estimating the probability density of a random variable $x$ from a set of predictive variables. Specifically, we need to construct a function $f_x$ that maps the values of the predictive variables $\mathbf{y} = (y_1, \ldots, y_N)$ into the space of probability densities over the possible values of $x$, such that
\begin{equation}
   f_x(\mathbf{y}) \approx  p(x|\mathbf{y}) ~.
  \label{eq: conditional density estimation} 
\end{equation}
In machine learning, the function $f_x$ is usually part of a large parametric family whose parameters have to be learned from a training set consisting of many $(x, \mathbf{y})$ pairs. In the following we will review some well known methods for CDE.

\subsection{Mixture density and quantized softmax networks}
A possible approach is to model $f_x$ as a convex mixture of simple parameterized probability densities in which both the mixing coefficients and the parameters are determined by the output of a DNN \cite{bishop1994mixture}. While the mixture density network has the advantage of exploiting the continuous  nature of the random variable, it enforces very strong constraints on the form of the resulting conditional density. A more flexible alternative is to quantize the range of $x$ into a finite number of intervals whose probability can be modeled independently using a softmax distribution~\cite{oord2016pixel}. The resulting probability density is a piecewise constant function:
\begin{equation}
   f_x(\mathbf{y}) = \frac{\sum_j \mathfrak{U}_j(x) e^{z_j\left(\mathbf{y}; W\right)}}{\sum_j e^{z_j\left(\mathbf{y}; W\right)}}  ~,
  \label{eq: quantized softmax network} 
\end{equation}
where $z_j\left(\mathbf{y}; W\right)$ denotes the activation of the $j$-th unit of the outer layer of a DNN (parametrized by a set of weights $W$) and  $ \mathfrak{U}_j(x)$ is the uniform probability density function on the $j$-th interval. 

Given a training set of $Q$ independently sampled training pairs $\{(x^{(q)},\mathbf{y}^{(q)}) \}$, the weights of the neural networks can be estimated by minimizing the negative log likelihood:
\begin{equation}
   C(W) = - \sum_q^Q \log f_{x^{(q)}}\left(\mathbf{y}^{(q)}\right)~,
  \label{eq: mixture log likelihood} 
\end{equation}
where the index $q$ denotes the $q$-th training pair. 

\subsection{Kernel density estimation}
Kernel density estimation is one of the oldest and best known methods for estimating a probability density $p(x)$ from a series of samples $x^{(1)},\ldots,x^{(P)}$ without relying on a parametric model \cite{rosenblatt1956remarks,parzen1962estimation}. The estimation is performed by centering a symmetric and normalized kernel function on each of the sample points:
\begin{equation}
   p(x) \approx \frac{1}{\sum_p w_p}\sum_p w_p \mathcal{K}\left(x,x^{(p)}\right)
  \label{eq: kernel density estimation}~,
\end{equation}
where the positive-valued weights $w_p$ determine the importance of each sample point. The kernel function $\mathcal{K}\left(x,x^{(p)}\right)$ is usually chosen a priori. A common choice is the Gaussian kernel
\begin{equation}
  \mathcal{K}_G(x,x'; \sigma) = \frac{1}{\sqrt{2 \pi} \sigma} e^{-\frac{\left(x - x'\right)^2}{2 \sigma^2}}
  \label{eq: Gaussian kernel}~,
\end{equation}
where the parameter $\sigma$, usually referred to as bandwidth, regulates the width of the kernel. The model parameters, weights and bandwidth can be optimized using frequentist or Bayesian methods \cite{turlach1993bandwidth, zhang2006bayesian}. In order to avoid the bandwidth selection problem, we reframe the estimation as follows:
\begin{equation}
   p(x) \approx \frac{1}{\sum_{p, j} w_{pj}}\sum_{p,j} w_{pj} \mathcal{K}_G\left(x,x^{(p)}; \sigma_j\right)
  \label{eq: generalized kernel density estimation}~,
\end{equation}
where, instead of optimizing the bandwidth parameters, we keep them fixed and extend the model by placing a whole family of kernel functions on each sample point. We refer to this method as \emph{kernel mixture density estimation}. More generally, we can apply kernel mixture density estimation for any given family of kernel functions $\{ \mathcal{K}_j(x,x') \}$. These functions can be defined on sets other than $\mathds{R} \times \mathds{R}$. For example, the following von Mises kernel can be used to estimate densities on a circle:
\begin{equation}
  \mathcal{K}_{VM}(\vartheta,\vartheta'; \kappa) = \frac{1}{2 \pi I_0(\kappa)} e^{-\kappa \cos(\vartheta - \vartheta')}
  \label{eq: VM kernel}~,
\end{equation}
where $I_0(x)$ is the modified Bessel function of order $0$.

Traditional methods for estimating conditional densities using kernel density estimation involve the independent estimation of joint and marginal densities, respectively $p(x, \mathbf{y})$ and $p(\mathbf{y})$~\cite{de2003conditional}. Unfortunately, in machine learning applications, this approach is often infeasible as it requires the proper estimation of probability densities in high-dimensional spaces.

\section{The kernel mixture network}
The main idea of this paper is to generalize the quantized softmax network using the kernel mixture density estimation approach. From Eq.~\ref{eq: quantized softmax network} we can see that the conditional density function of a quantized softmax network is a special case of the expression in Eq.~\ref{eq: kernel density estimation} where the kernels are the rectangular functions $\mathfrak{U}_j(x)$ and the weights are given by $e^{z_j}$. It is therefore natural to extend this expression to a more general family of kernel functions that can exploit the topology of continuous random variables. We introduce the KMN by using a model of the form given in Eq.~\ref{eq: generalized kernel density estimation} in which the weights are determined by the output of a DNN:
\begin{equation}
   f_x(\mathbf{y}) = \frac{1}{\sum_{p, j} w_{pj}\left(\mathbf{y}; W\right)}\sum_{p,j} w_{pj}\left(\mathbf{y}; W\right) \mathcal{K}_j\left(x,x^{(p)}\right)~.
  \label{eq: kernel density estimation network}
\end{equation}
In order to assure that all weights are non negative we assume that all the output nodes of the networks have non-negative activation functions.

The set of center points $\left\{x^{(p)}\right\}$ is composed of all the values assumed by the variable $x$ in the training set. Consequently the variable range does not need to be known a priori. In other words, the dimensionality of the functional space spanned by the KMN depends on the number of training points. This feature is shared with other nonparametric regression methods such as Gaussian process regression~\cite{rasmussen2006}. However, using as many kernels as (a multiple of the) training points can be impractical in big datasets. Therefore we subsample $\left\{x^{(p)} \right\}$ by recursively removing each center point that is closer than a constant $\delta$ to its predecessor. 

The loss function of the kernel density estimation network can be obtained by plugging the model given in Eq.~\ref{eq: kernel density estimation network} into the expression for the negative log likelihood given in Eq.~\ref{eq: mixture log likelihood}:
\begin{equation}
   C(W) = - \sum_q^Q \left[ \log{\sum_{p,j} w_{pj}\bigl(\mathbf{y}^{(q)}; W) \mathcal{K}_j(x^{(q)},x^{(p)}\bigr)} - \log{\sum_{p,j} w_{pj}\bigl(\mathbf{y}^{(q)}; W\bigr)}\right]
  \label{eq: kernel mixture cost function}.
\end{equation}
This cost function can be optimized with respect to the neural network parameters using standard back-propagation techniques. 

Note that we have complete freedom in choosing the kernel functions as far as they define valid probability densities. This property can be used for estimating conditional densities on arbitrary manifolds or even on discrete objects such as graphs. Importantly, the kernel itself does not need to be differentiable since the gradients are computed with respect to the weights.

\section{Experiments}
In this section we validate the performance of the KMN on a Bayesian filtering problem and on a generative modeling problem. We compare the performance of the kernel mixture approximate Bayesian filter with both its quantized analog and the extended Kalman filter (EKF), the most widely used method for nonlinear filtering~\cite{sorenson1985kalman}. In the generative modeling application we use the LSTM-PCA approach for generating grayscale and color images of human faces.

\subsection{Applications to Bayesian filtering}
Bayesian filtering is a special case of Bayesian inference where the current state $x_t$ of a latent time series has to be estimated from a series of past indirect measurements $y_1,...,y_{t-1}$~\cite{sarkka2013bayesian}. The structure of a filtering problem allows for a convenient recursive reformulation in terms of the Bayesian filtering equations~\cite{sarkka2013bayesian}. Unfortunately these equations are very often intractable. In these cases, the solution has to be approximated using methods such as the EKF. Here, we introduce the use of KMN to estimate the density $p(x_t | y_1,...,y_{t-1})$ from a large set of simulated samples drawn from the prior distribution. Specifically, we sample latent time series from the prior distribution and, subsequently, synthetic observations from the likelihood. This approach is a probabilistic extension of our recently introduced ConvNet smoother \cite{ambrogioni2017estimating} and is an application of the general framework of \textepsilon-free approximate Bayesian inference~\cite{papamakarios2016fast}.

In our first simulation, we generated a latent time series by integrating the following stochastic oscillator equation:
\begin{equation}
\frac{d^2x}{dt^2} = -\omega_0^2 x - \beta \frac{dx}{dt} + k_2 x^2 + k_3 x^3 ~ + \xi(t)\,.
\label{eq: nonlinear SDE}
\end{equation}
The stochastic dynamical process was discretized using the Euler-Maruyama scheme with integration step equal to $0.01$ seconds. The parameters of the dynamical model were: $\omega_0 = 5$, $\beta = 0.2$, $k_2 = 15$ and $k_3 = -0.5$. We simulated $499800$ training time series and $200$ validation time series. Noisy observations were generated by adding Gaussian white noise to the time series. We trained a deep CNN to estimate the probability of the latent state given the past noisy observations. The details of the architecture are the same as used in \cite{ambrogioni2017estimating}. The only difference is that here the output layer determined the weights of the kernel mixture. We used rectified quadratic units as activation functions on the final layer. As kernels we used Gaussian functions with standard deviations ranging from $0.25$ to $2.75$ in steps of $0.5$.
 
 \begin{figure}[!ht]
	\centering
   	\includegraphics[width=1.\textwidth] {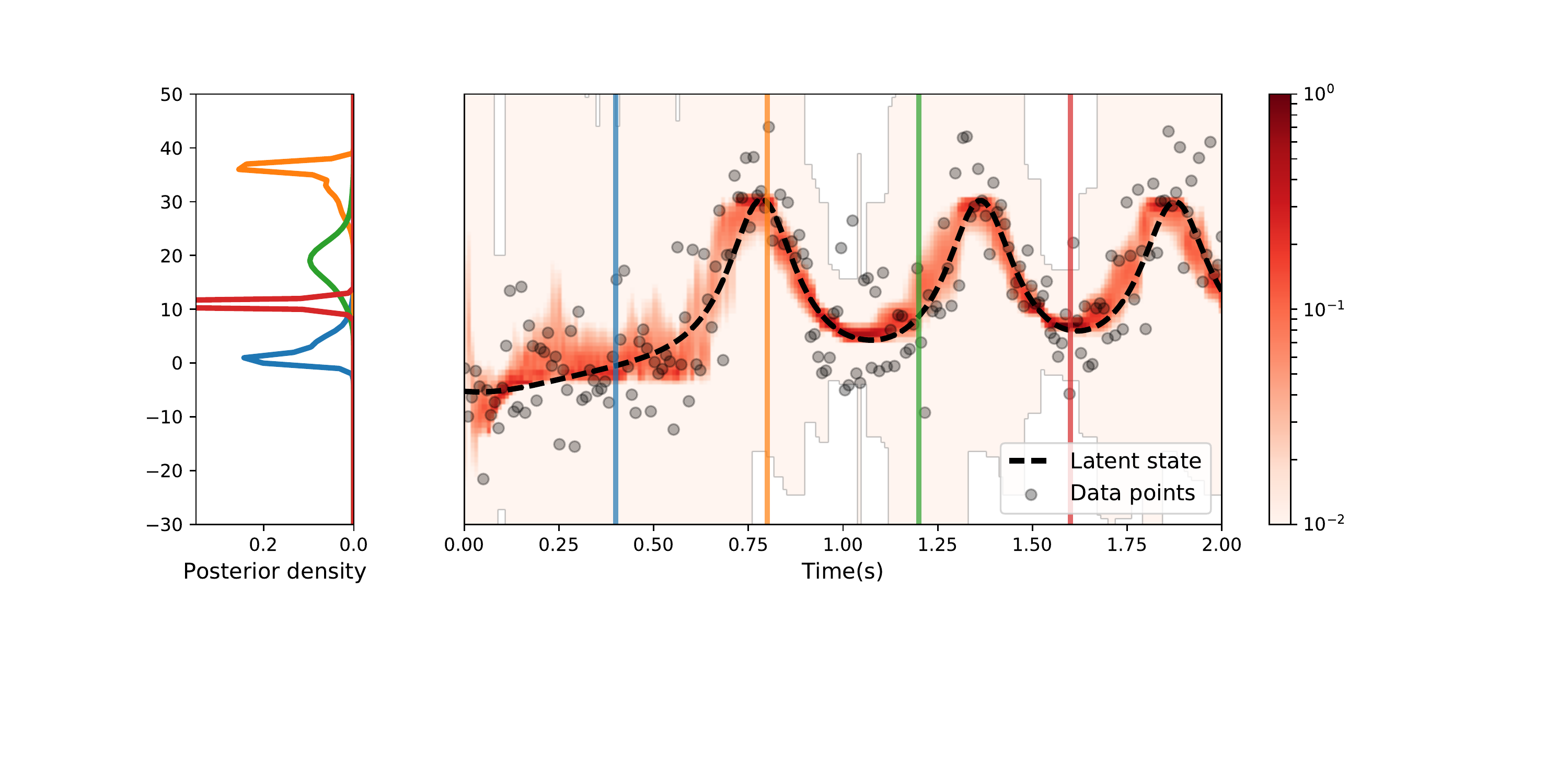}
	\caption{Approximate posterior densities of the state of a nonlinear oscillator obtained using the KMN Bayesian filter with Gaussian kernels. The red color on the right panel represents the probability density of the filter given all the past observations (gray dots). The dashed line is the underlying ground truth signal. The colored vertical lines mark the position of four time points. The conditionals at these points are plotted on the left panel in a color coded manner. }
	\label{figure 1}
\end{figure}

Figure~\ref{figure 1} shows the resulting marginal posterior filter densities of an example trial. As we would expect from a filtering problem, the posterior density becomes tighter as time progresses since the filter can use increasingly more data. Importantly, the KMN can model very non-Gaussian distributions, in this case recovering the skewness of the posterior distribution. 

\begin{figure}[!ht]
	\centering
    	\includegraphics[width=1.\textwidth] {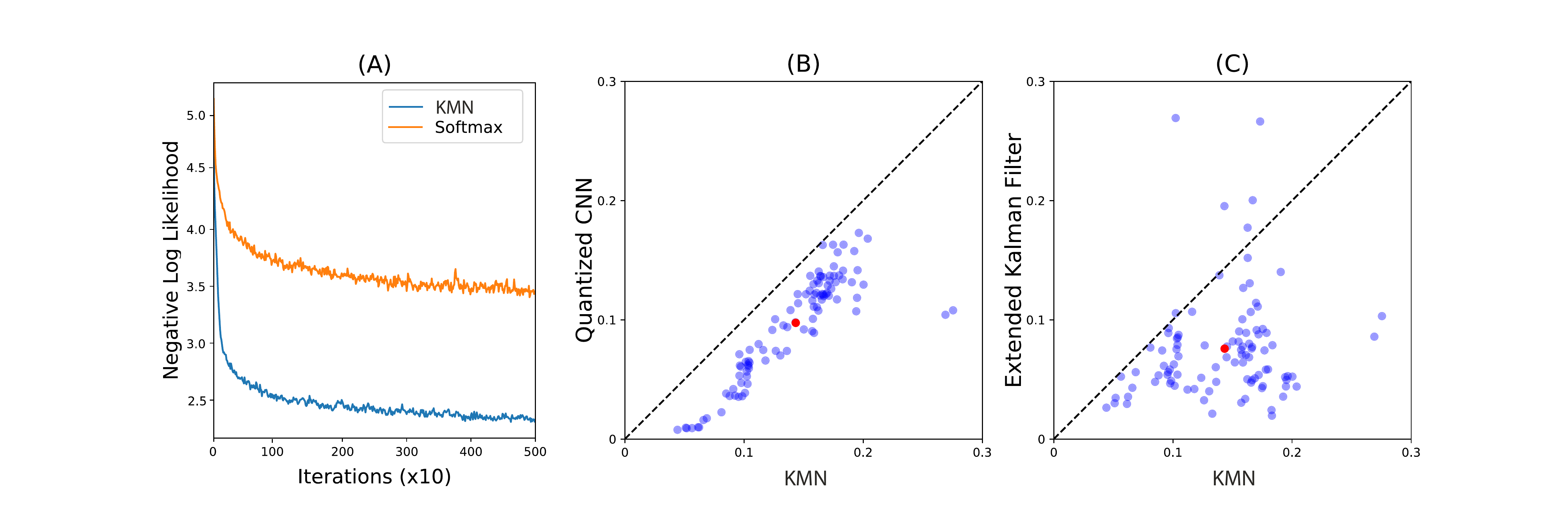}
	\caption{Comparison between filtering methods on the analysis of a nonlinear oscillator. A) Validation loss (negative log likelihood) of quantized and kernel CNN as function of training iteration. B) Comparison between kernel CNN and quantized CNN. The blue dots are individual validation trials and the red dot is the center of mass. C) Comparison between kernel CNN and EKF.}
	\label{figure 2}
\end{figure}

In Fig.~\ref{figure 2} we compare the performance of the KMN filter with the quantized CNN filter (same architecture, bin size equal to $0.25$) and the more conventional EKF. Panel A shows the validation set loss (negative log likelihood) of the kernel mixture and the quantized networks as a function of the number of training iterations. The KMN converges faster than its quantized alternative and reaches a better local minimum. The scatter plots in panels B and C show the comparisons of the likelihood of the trained KMN with the likelihood of respectively the trained quantized CNN and the EKF. Our KMN outperforms both alternative methods in almost all the validation trials.

In our second simulation we give an example of KMN conditional density estimation on a manifold other than $\mathds{R}$. In particular, we use the KMN Bayesian filter for estimating the phase of an an-harmonic wave with random nonlinear waveform from noisy measurements. The phase is a circular variable that can be parametrized with an angle $\vartheta$. Our latent state was a phase with uniform random initial value $\vartheta_0$ and fixed linear growth: $\vartheta(t) = \vartheta_0 + 4 \pi t$. The indirect measurements were obtained as follows:
\begin{equation}
y_t = f\big(\cos{\vartheta(t)}\big) + w_t~,
\label{eq: phase measurements}
\end{equation}
were $w_t$ is Gaussian white noise (sd = 2). The random function $f(x)$ is given by
\begin{equation}
f(a) = w_1 a + w_2 a^2 + w_3 a^3 + w_4 a^4 + w_5 a^5~,
\label{eq: random nonlinearity}
\end{equation}
where the Taylor coefficients $w_1,w_3$ and $w_5$ were sampled from truncated t distributions (df = 3, from $0$ to $\infty$) and the coefficients $w_2$ and $w_4$ were sampled from t distributions (df = 3). Note that the likelihood function of $y_t$ would be very challenging to obtain in closed form, however this is not a problem for our approach (see \cite{ambrogioni2017estimating} for more details). In order to exploit the circularity of the phase variable we used the von Mises kernels given in Eq.~\ref{eq: VM kernel}, with scale $1/\sqrt{\kappa}$ ranging from $\pi/250$ to $2 \pi /25$ in steps of $\pi/250$. 

\begin{figure}[!ht]
	\centering
    	\includegraphics[width=1.\textwidth] {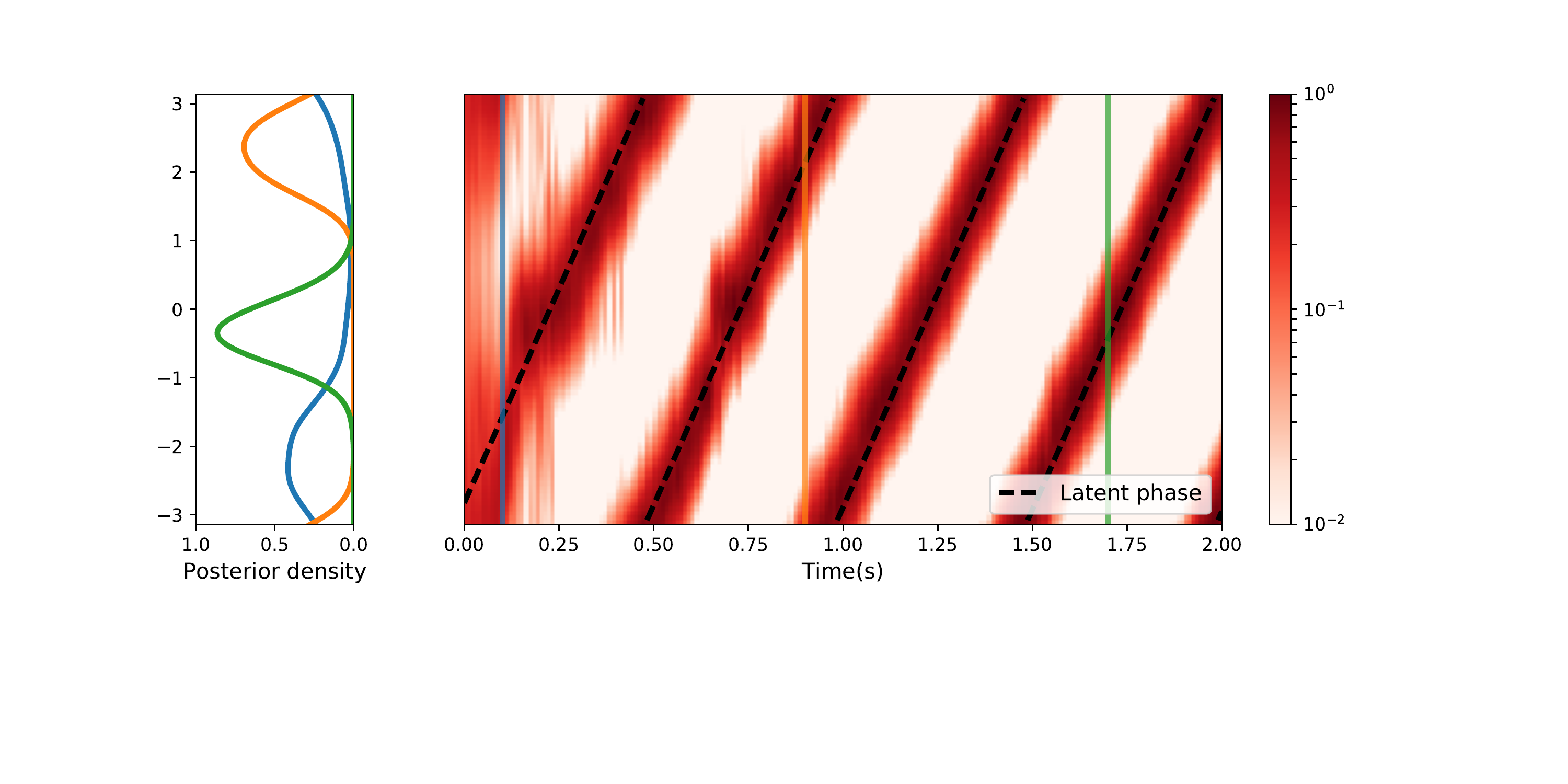}
	\caption{Approximate posterior densities of the phase of an an-harmonic wave obtained using the KMN Bayesian filter with von Mises kernels.}
	\label{figure 3}
\end{figure}

Figure~\ref{figure 3} shows the posterior filter density for an example trial. As expected, the filter is initially very uncertain, however it quickly converges to accurately track the underlying phase. Note that the resulting conditional densities are defined over a circle, since the phase $-\pi$ is equivalent to the phase $\pi$.

\subsection{Applications to generative modeling}
We performed two face generation experiments to compare the performance of the KMN with a quantized softmax approach. In the first experiment we trained two networks to address the problem of grayscale face generation. In the second experiment we tackled the more complex case of colored face generation. The networks shared the same LSTM-PCA architecture, only differing in their last layers. Specifically, they had two 512-unit LSTM layers followed by either a rectified linear layer (in the case of KMN) or a softmax layer (in the case of quantized softmax).

We first performed PCA on the aligned face images in the CelebA dataset after resizing them to $64 \times 64$ pixels and determined the principal components of the faces explaining 90\% of the variance in the data, corresponding to 95 and 125 components in the cases of grayscale faces and color faces respectively. The task of the networks was to predict the loading of each principal component given all the loadings of the higher variance components. For the softmax model we quantized the components to 256 bins with equal widths. We used the training split of the CelebA dataset to train the models, the test split to test them and the validation split to monitor the loss during training. We trained all the models with the Adam optimizer~\cite{adam} for 100 epochs.

\begin{figure}[!ht]
	\centering
    	\includegraphics[width=1.\textwidth] {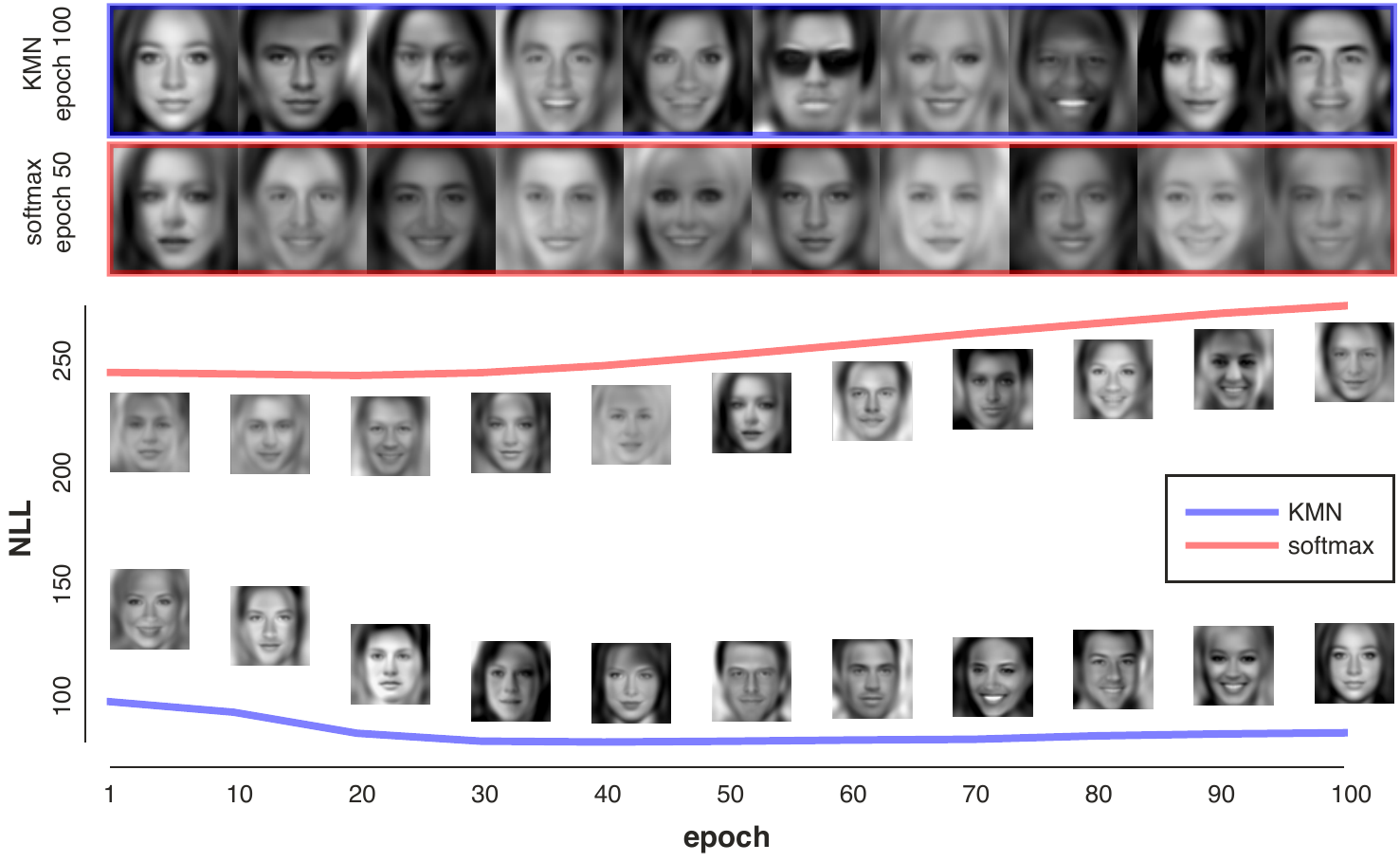}
	\caption{Grayscale faces generated by KMN and softmax models with LSTM-PCA architecture (top panel) and the progress of the test losses of both models during training (bottom panel).}
	\label{figure 4}
\end{figure}

\begin{figure}[!ht]
	\centering
    	\includegraphics[width=1.\textwidth] {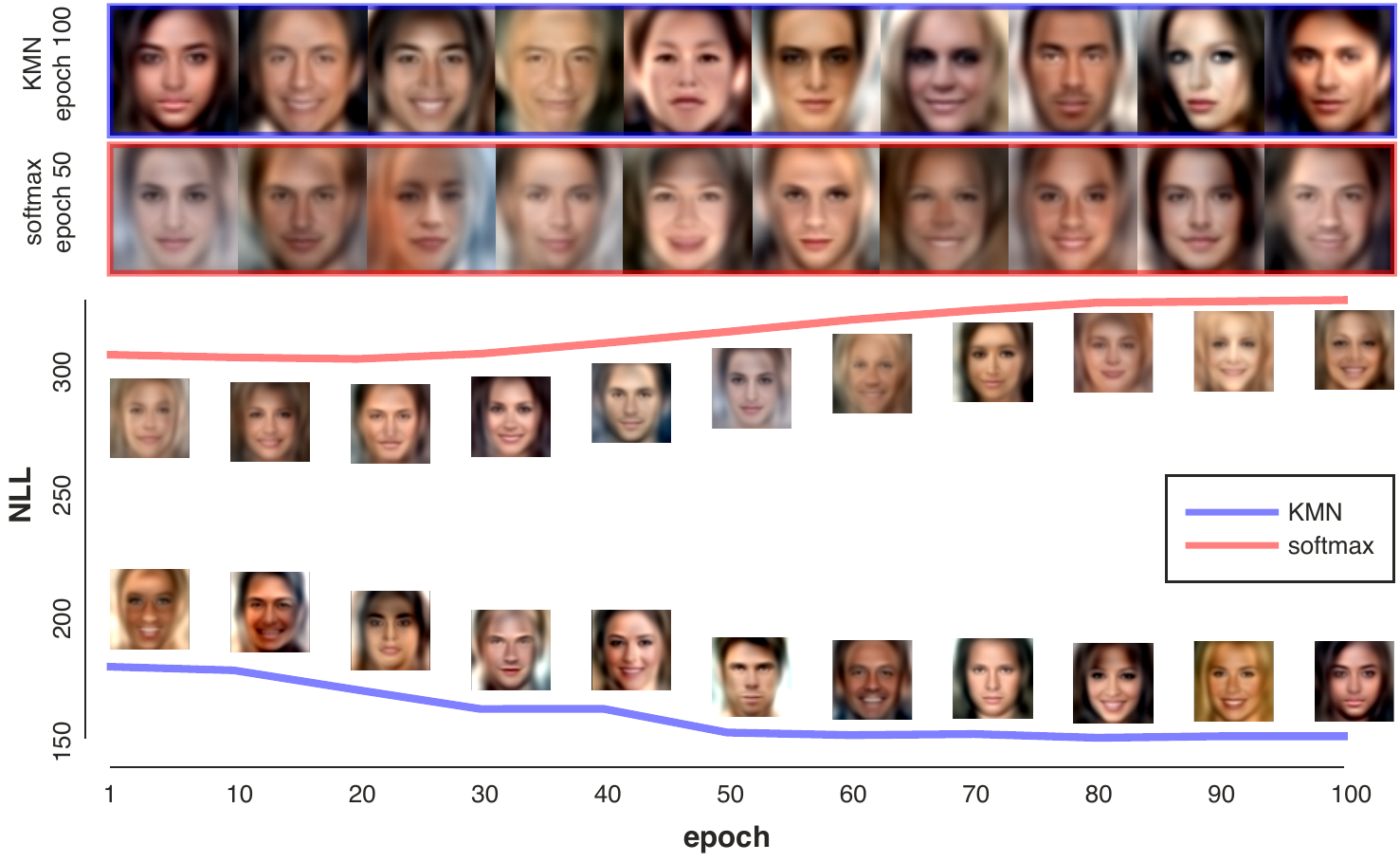}
	\caption{Color faces generated by KMN and softmax models with LSTM-PCA architecture (top panel) and the progress of the test losses of both models during training (bottom panel).}
	\label{figure 5}
\end{figure}

In both the grayscale and the color face generation experiments, the validation loss during training indicates that the softmax models overfitted the training set after approximately 50 epochs, whereas the KMN models seemed to continue learning without overfitting by the 100th epoch. Furthermore, we observed large differences between the negative log likelihoods of the softmax model and the KMN model on the test set in both experiments in favor of the KMN model (Figures~\ref{figure 4} and \ref{figure 5}, bottom panel). In order to further evaluate the performance of the two methods we generated face images (from epoch 50 for softmax, i.e. before overfitting, and from epoch 100 for KMN). Both the grayscale and the color faces generated by the KMN models appeared more realistic. Specifically, they were sharper and less blurry and had fewer artifacts compared to those generated by the softmax models (Figures~\ref{figure 4} and \ref{figure 5}, top panel). Furthermore, the KMN-generated faces were visibly more diverse than those generated by the softmax models.

\section{Discussion}
We introduced a new method for the nonparametric estimation of conditional probability density functions using neural networks. The KMN combines the flexibility of the popular quantized softmax approach with the regularizing properties of kernel density estimation methods. We showed that the KMN can be used for constructing Bayesian filters that track very complex probability densities on manifolds. Furthermore, we used the KMN for generating images of human faces using the newly introduced LSTM-PCA network architecture. We showed that, given the same architecture, the KMN network approach is less likely to overfit than the quantized softmax approach, generating images that are more realistic and more diversified.

Note that the KMN can be used together with several other popular generative methods, such as PixelCNN~\cite{van2016conditional}, PixelRNN~\cite{pixelrnn} and WaveNet~\cite{oord2016wavenet}. From our simulations, it is likely that the KMN approach will substantially improve the quality and diversity of the samples generated using these methods. The KMN approach may also be used together with models that estimate probability distributions over discrete graphs. This can be done by using diffusion kernels over the graph that encode the average distance between nodes \cite{kondor2002diffusion}. For example, image recognition techniques can exploit the semantic similarities between classes as provided by lexical databases such as WordNet~\cite{miller1995wordnet}. Also note that, while we have been using the KMN solely for density estimation, the method can easily be used for reconstructing arbitrary scalar functions on manifolds simply by using a different loss function, such as the mean squared loss. 

\medskip
\small

\bibliography{KernelDensity}
\end{document}